\documentclass{article}

\usepackage[final]{./styles/neurips_2019}

\usepackage[utf8]{inputenc} 
\usepackage[T1]{fontenc}    
\usepackage{hyperref}       
\usepackage{url}            
\usepackage{booktabs}       
\usepackage{amsfonts}       
\usepackage{nicefrac}       
\usepackage{microtype}      
\usepackage{algorithm}
\usepackage{algorithmic}
\usepackage{graphicx}
\usepackage{array}
\usepackage{caption}
\usepackage{subcaption}

\begin{document}

\title{Learning Digital Circuits: A Journey Through Weight Invariant Self-Pruning Neural Networks}

\author{Amey Agrawal, Rohit Karlupia\\
Qubole India\\
{\tt\small \{ameya, rohitk\}@qubole.com}}

\maketitle


\begin{abstract}

Recently, in the paper ``Weight Agnostic Neural Networks'' Gaier \& Ha utilized architecture search to find networks where the topology completely encodes the knowledge. However, architecture search in topology space is expensive. We use the existing framework of binarized networks to find performant topologies by constraining the weights to be either, zero or one. We show that such topologies achieve performance similar to standard networks while pruning more than 99\% weights. We further demonstrate that these topologies can perform tasks using constant weights without any explicit tuning. Finally, we discover that in our setup each neuron acts like a NOR gate, virtually learning a digital circuit. We demonstrate the efficacy of our approach on computer vision datasets.

\end{abstract}
\section{Introduction}

\citet{lottery} recently introduced the hypothesis that dense feed-forward networks contain a smaller sub-network, \textit{``The Winning Ticket''} which when trained in isolation reaches test accuracy comparable to the original network. They attribute the efficacy of this sub-network to an \textit{``Initialization Lottery''}, the specific set of initial weights that make the training effective. They show that re-initializing the weights of the winning ticket depletes the performance. 

Based on further analysis of the Lottery Ticket algorithm, \citet{uber} hypothesize that the lottery ticket algorithm works well only when the pruned weights were already headed to zero by gradient descent. They also demonstrate that only the signs initial weights are important and not their magnitude. They formulate an algorithm to identify \textit{``Supermask''} which represents a certain topology within the network that can produce reasonable results without any training just using random constant weights. However, the constant weights must have the same signs as the original initialization.

In another recent work, \citet{ha} discover that sparse network topologies can be learned via architecture search. Unlike \citet{uber}, they use a constant weight with same signs for all connections within the network. They demonstrate that the performance of these network topologies remains largely invariant to the choice of the weight. This naturally raises the question of whether such topologies can be learned with backpropagation.

In order to obtain fast inference on low-powered devices, \citet{bincon} proposed a framework, \textit{``BinaryConnect''}, which allows learning networks with weights are constrained to two possible values -1 and 1. In this paper, we modify this framework to learn network topology by constraining the weights to 0 and 1. We propose an explanation for why these topologies work by drawing a parallel to logic gates and successively introduce certain constructs that allow us to learn networks where each neuron acts like an OR gate and normalization layers mimic NOT gate. We further demonstrate that the topologies learned through this framework are weight agnostic and can solve complicated vision tasks.

\section{Self-Pruning Networks}
\label{sec:self_prun}

In BinaryConnect \citep{bincon}, binary weights are used during forward and backward propagation, however, the updates are performed on real-valued weights ($w$). The real-valued weights are binarized at the beginning of forward pass. Having high precision weights allows gradient descent to make many small updates which subsequently change the the binarized values. Constraining the weights to values zero and one allows us to learn sparse network topologies. In this section, we describe the changes which allow BinaryConnect framework to work with our modified constraints.

\subsection{Binarization \& Weight Clipping}

\citet{bincon} obtain binarized weights ($w_b$) using the sign function. As a natural extension, we use a simple step function for binarization:

\begin{equation}
\label{eq:bin}
    w_b = \left\{ \begin{array}{ll}
            1 & \mbox{if $w \geq 0.5$},\\
            0 & \mbox{otherwise}.\end{array} \right.
\end{equation}

In all our experiments, we also binarize our inputs as described in section \ref{sec:exp}. For the sake of simplicity we do not use bias in our neurons. Furthermore, the BinaryConnect model employs weight clipping, a widely used regularization technique. Since in our setting we binarize weights to \{0, 1\}, we change the clipping range to [0, 1] from the the standard range [-1, 1] used in BinaryConnect.

\subsection{Activation Function \& Normalization}

Similar to the original BinaryConnect network we employ Batch Normalization (BN) \citep{ioffe2015batch}. However, we use $\tanh$ instead of Rectifier Linear Units (ReLU) \citep{nair2010rectified} as activation function. We further discuss these choices in section \ref{sec:decon}.

\subsection{Weight Initialization}

Most standard weight initialization techniques use a normal distribution with zero mean. However, due to our binarization scheme shown in Eq. \ref{eq:bin}, most weights drawn from such a distribution would be binarized to zero. We would want to initialize our weights from a bimodal distribution such that post binarization, we have both weights set to zero and one. Hence, we use a Bernoulli distribution for weight initialization. In our experiments we find that initialization with success probability ($p$) anywhere in the range [0.0001, 0.04] performs fairly well.

\begin{algorithm}[H]
\begin{algorithmic}
    \REQUIRE a minibatch of (inputs, targets),
    previous parameters $w_{t-1}$ (weights).
    \ENSURE updated parameters $w_t$.  
    \STATE {\bf 1. Forward propagation:}
    \STATE $w_b \leftarrow {\rm binarize}(w_{t-1})$
    \STATE For $k=1$ to $L$, compute activactions $a_k$ knowing $a_{k-1}$ and $w_b$
    \STATE {\bf 2. Backward propagation:}
    \STATE Initialize output layer's activations gradient $\frac{\partial C}{\partial a_L}$
    \STATE For $k=L$ to $2$, compute $\frac{\partial C}{\partial a_{k-1}}$
          knowing $\frac{\partial C}{\partial a_k}$ and $w_b$
    \STATE {\bf 3. Parameter update:}
    \STATE Compute $\frac{\partial C}{\partial w_b}$ 
        knowing $\frac{\partial C}{\partial a_k}$ and $a_{k-1}$
    \STATE $w_t \leftarrow {\rm clip}(w_{t-1} - \eta \frac{\partial C}{\partial w_b})$
\end{algorithmic}
\caption{SGD training with BinaryConnect. $C$ is the cost function for minibatch and the functions binarize($w$) and clip($w$) specify how to binarize and clip weights. $L$ is the number of layers.
}
\end{algorithm}
\section{Learning Digital Circuits}
\label{sec:decon}

We obtain near state-of-the-art results on MNIST using the framework described in section \ref{sec:self_prun}. However, we make a peculiar observation that if we remove the Batch Normalization layers from our architecture, the network performs only marginally better than random baseline. In rest of this section, we deconstruct the network to understand why Batch Normalization is critical to the the performance of this network and propose alternative components which allow us to learn networks which virtually act like digital circuits.

\subsection{Network without Batch Normalization}

As a direct consequence of constraining all the weights to zero and one, each neuron acts as a simple summation of all the inputs from previous layer. Since we have also binarized inputs, output of each neuron in the input layer would be a non-negative integer. When we apply $\tanh$ activation on these values, the output would be semi-binary, that is either 0 or close to 1 ($\tanh(2) = 0.96$), acting like a logical OR gate. Because of the semi-binary nature of the output of this layer, we can extend the same reasoning to all the subsequent layers in the network. Hence, performing gradient descent to learn the topology in this network is similar to finding a circuit built only with OR gates which solves the task. OR gate is not an universal gate and thus is limited in its expressibility, which could be why our network does not work without Batch Normalization.

\subsection{Leveraging Batch Level Statistics}

Let us now consider a batch of inputs received by the Batch Normalization layer in our network. If the mean of inputs is close to 1, the high inputs (1s) would be normalized to low signals and vice versa. In this scenario, the Batch Normalization layer would act as a NOT gate. On the other hand, if the mean of the input batch is close to 0, the layer would preserve the polarity of the signals. Based on the batch statistics the output thus, either represents a logical NOR or OR gate. Being an universal gate NOR gate should provide us the expressibility required to solve complicated problems.

\subsection{Substituting Batch Normalization}

In order to evaluate the impact of the logical NOT like behaviour of Batch Normalization layer in isolation with its other properties, we replace the normalization layer with simple negation operation:

\begin{equation}
    Hard Negation(x) = 1 - x 
\end{equation}

As reported in subsection \ref{sec:exp} the network performs reasonably well when trained with hard negation layer. We also define a soft negation function as described below,

\begin{equation}
    Soft Negation(x) = x * (1 - \alpha) + (1 - x) * \alpha
\end{equation}

Here, invert gate ($\alpha$) is a scalar which we learn with back propagation. The values of $\alpha$ are clipped within the range [0, 1]. In our experiments soft negation works significantly better than hard negation. We also make an interesting observation that the learned values of $\alpha$ are either 0 or 1 and no values in between.

\section{Experiments}
\label{sec:exp}

\begin{figure}
\centering
\begin{subfigure}{.5\textwidth}
    \centering
    \includegraphics[width=\linewidth]{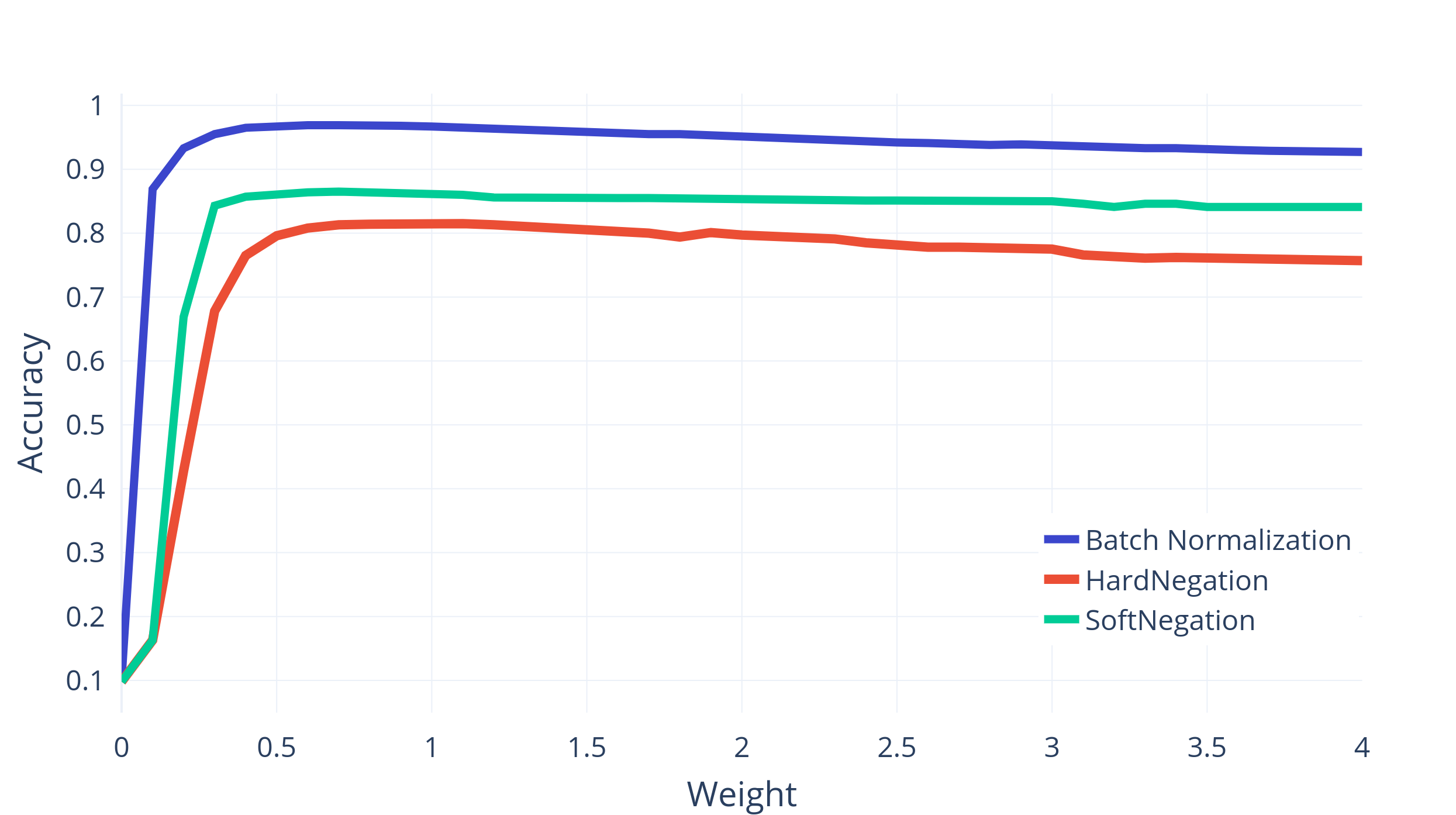}
    \caption{MNIST}.
    \label{fig:mnist_we}
\end{subfigure}%
\begin{subfigure}{.5\textwidth}
  \centering
    \includegraphics[width=\linewidth]{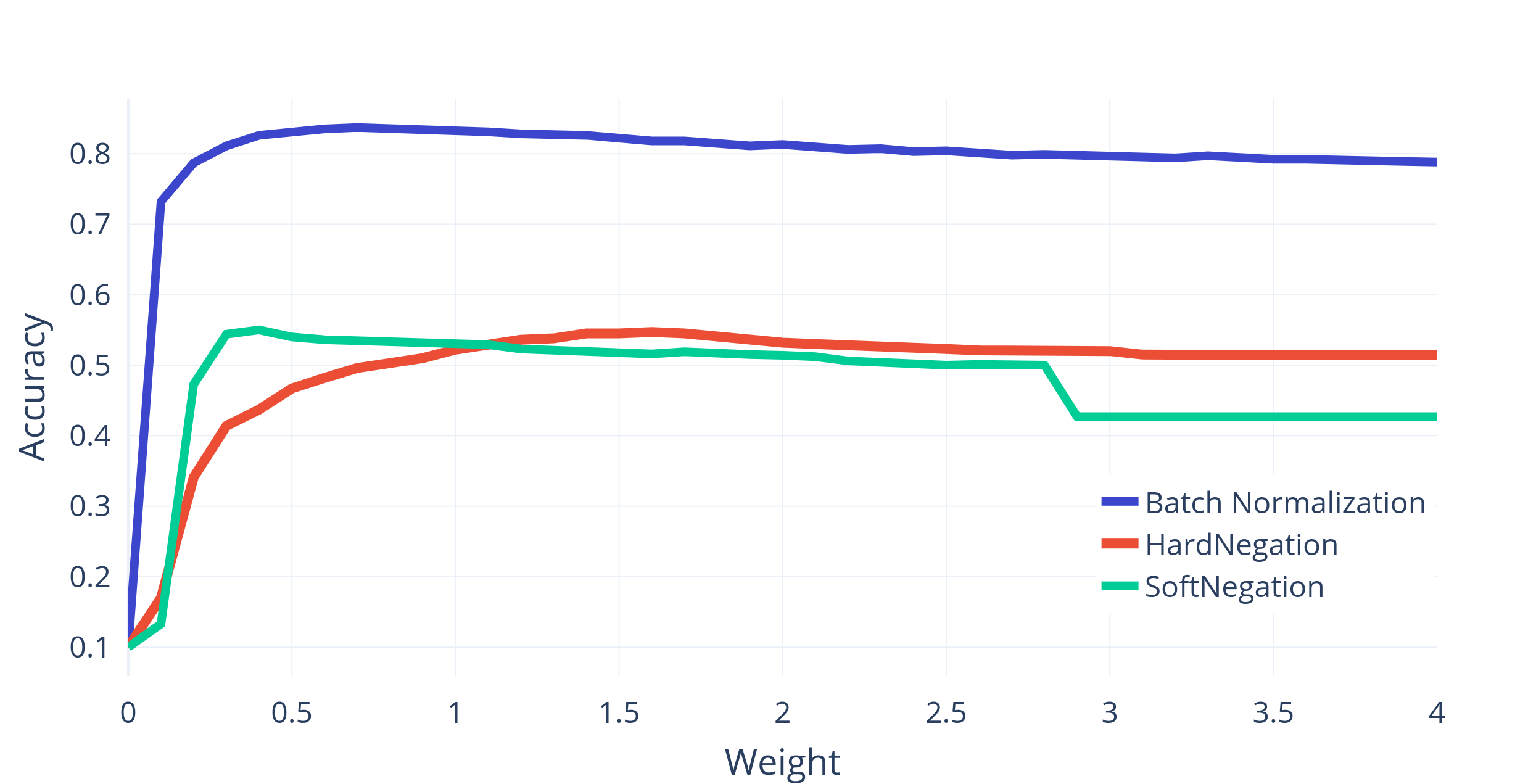}
    \caption{Fashion-MNIST}.
    \label{fig:fashion_we}
\end{subfigure}
\caption{Variation in test accuracy with weight in Self-Pruning Networks.}
\end{figure}

In this section, we show the performance of our framework on MNIST and Fashion-MNIST datasets. We also demonstrate that the performance of the learned network topologies is largely invariant to the weight. 

\begin{table}[htbp]
  \caption{Test accuracy of networks trained on the MNIST and Fashion-MNIST.}
\label{tbl:accuracy}
  \centering
  \begin{tabular}{lll}
    \toprule
    Method     & MNIST     & Fashion-MNIST \\
    \midrule
    Real-Valued Network                           & 98.1\%            & 89.5\%  \\
    \midrule
    Self-Pruning Network (BN)                     & 96.7\%            & 83.2\%  \\
    Self-Pruning Network (Hard Negation)           & 81.5\%            & 52.9\%  \\
    Self-Pruning Network (Soft Negation)           & 86.0\%            & 53.3\%  \\
    WANN (Tuned Weight) \citep{ha}                 & 91.9\%            & -       \\
    Supermask (Signed Constant) \citep{uber}       & 86.3\%            & -       \\
    \bottomrule
  \end{tabular}
\end{table}

\subsection{MNIST}

MNIST \citep{lecun1998mnist} is standard image classification benchmark dataset with 28 x 28 gray-scale images of handwritten digits. We binarize the images using Eq. \ref{eq:bin}. Our model consists of three dense hidden layers with 2048 units each and $\tanh$ activation. We use negative log likelihood loss. As shown in Table \ref{tbl:accuracy}, we obtain near state-of-the-art performance with Batch Normalization. Experiments show that the soft negation operation performs significantly better than hard negation by allowing combinations of OR and NOR layers. We can see that greater than 99\% of the connections are pruned (set to 0) in Table \ref{tbl:pruned}.

\subsection{Fashion-MNIST}

Fashion-MNIST was proposed by \citet{xiao2017fashion} as a drop in replacement for MNSIT dataset with more complexity. It contains 28 x 28 grey-scale images of apparel with labels from 10 classes. We find that the binarized network with Batch Normalization achieve accuracy very close to an identical network with real-valued weights. However, we see a significant drop in the performance with the negation layers. 

\subsection{Weight Invariance}

Intriguingly, we observe that in our learned networks a large number of neurons output values greater than one. These outputs saturate to values close to one due to the $\tanh$ activation. Hence, the performance of the network should not be greatly affected if we re-scale the output of neurons by changing the weights. To verify this hypothesis, we test the performance of previously learned topologies on weights in the range [0, 4]. Additionally, we do not freeze Batch Normalization parameters to facilitate the adoption of new weights. Figures \ref{fig:mnist_we}, \ref{fig:fashion_we} show that indeed the performance of these topologies is largely invariant to the weights. 

\begin{figure*}[htbp]
\centerline{
    \includegraphics[width=20cm]{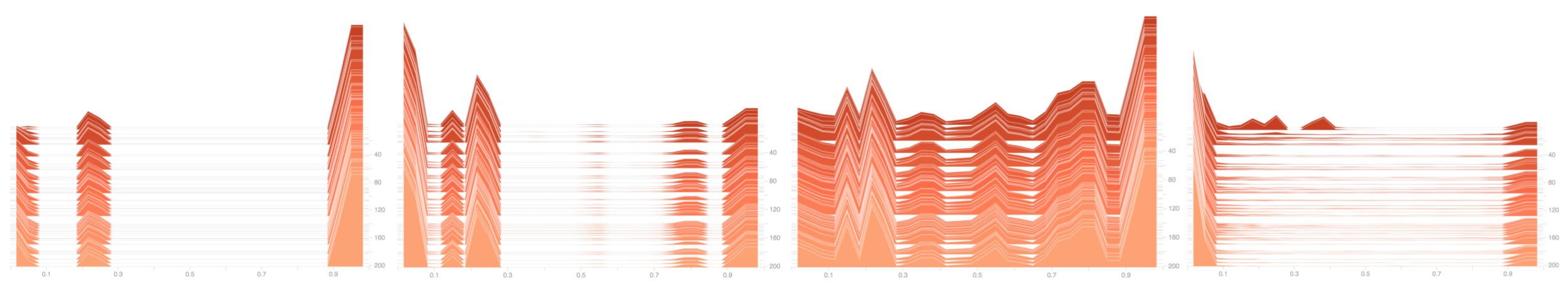}
}

\caption{Histograms show output of hard negation layer throughout training on MNIST dataset. Number of training batches is on z-axis.}
\label{fig:hard_histo}
\end{figure*}
\begin{figure*}[ht]
\centerline{
    \includegraphics[width=20cm]{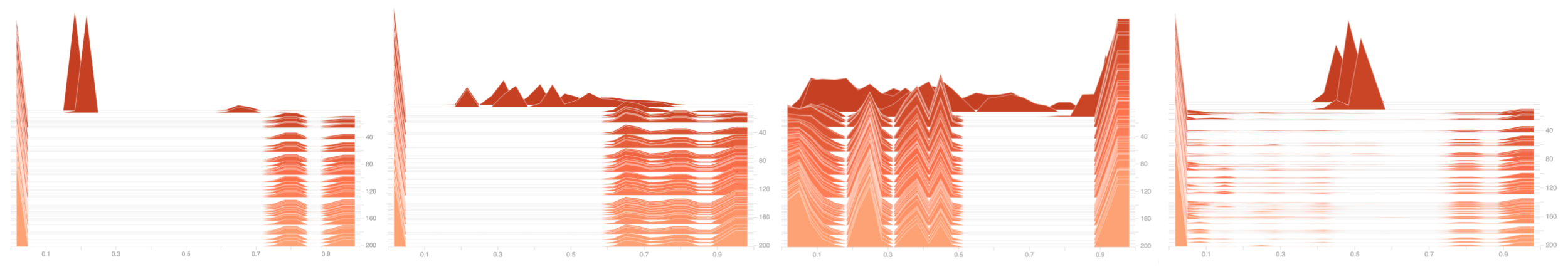}
}

\caption{Histograms depicting output of soft negation layer throughout training on MNIST dataset. z-axis represents the number of training batches. We can observe that the values of activations saturate towards zero and one as the training progresses and ($\alpha$) is learned.}
\label{fig:soft_histo}
\end{figure*}

\begin{table}[htbp]
  \caption{Percentage of pruned weights for each layer.}
  \label{tbl:pruned}
  \centering
  \begin{tabular}{lll}
    \toprule
    Normalization     & MNIST     & Fashion-MNIST \\
    \midrule
    Batch Normalization    & 99.36\%, 99.49\%, 99.33\%, 95.56\%        & 99.08\%, 99.49\%, 99.86\%, 97.16\%     \\
    Hard Negation          & 99.18\%, 99.59\%, 99.87\%, 99.30\%        & 99.24\%, 99.63\%, 99.74\%, 93.36\%     \\
    Soft Negation          & 98.97\%, 99.73\%, 99.89\%, 99.83\%        & 98.81\%, 99.69\%, 99.86\%, 99.78\%     \\
    \bottomrule
  \end{tabular}
\end{table}
\vspace{1 cm}

\section{Conclusion}

In this paper, we propose a method to augment the BinaryConnect \citep{bincon} framework to learn networks with weights zero and one, thus enabling the network to prune weights directly with gradient descent. We show that the topologies learned with our framework achieve comparable performance to their real-valued counterparts. We also demonstrate that these networks are weight agnostic in nature. We observe that Batch Normalization is critical to the functioning of our framework. In order to understand this phenomenon we deconstruct our architecture to discover the role batch-level statistics play in-order in the functioning of the network. We then propose negation layers that allow us to learn networks in which each neuron virtually act as a NOR gate.

{\small
\bibliographystyle{apalike}
\bibliography{main}
}

\end{document}